\documentclass[11pt,a4paper]{article}
\usepackage[hyperref]{emnlp-ijcnlp-2019}
\usepackage{times}
\usepackage{latexsym}

\usepackage{amsmath, amssymb, amsthm}
\usepackage{booktabs}
\usepackage{graphicx}
\usepackage{makecell}
\usepackage{hhline}
\usepackage[caption=false]{subfig}

\usepackage{xcolor}

\usepackage{pgfplotstable}
\usepackage{pgfplots}
\pgfplotsset{width=0.47\textwidth,compat=1.9, height=1.6in}

\usepackage{url}
\usepackage[normalem]{ulem}

\usepackage[symbol]{footmisc}

\definecolor{lblue}{HTML}{DAE8FC}
\definecolor{dblue}{HTML}{6C8EBF}
\definecolor{dgreen}{HTML}{33A02C}
\definecolor{dred}{HTML}{E31A1C}
\newcommand{\dblue}[1]{{\color{dblue} #1}}
\newcommand{\dgreen}[1]{{\color{dgreen} #1}}
\newcommand{\dred}[1]{{\color{dred} #1}}

\usepackage{url}

\aclfinalcopy 

\title{Towards Understanding of Medical Randomized Controlled Trials\\by Conclusion Generation}

\author{Alexander Te-Wei Shieh\footnotemark[1]\quad Yung-Sung Chuang\footnotemark[1]\quad Shang-Yu Su\quad Yun-Nung Chen\\
National Taiwan University, Taipei, Taiwan \\
  {\tt \{b05401009,b05901033,f05921117\}@ntu.edu.tw\quad y.v.chen@ieee.org } \\}

\date{}

\begin{document}
\maketitle
\footnotetext[1]{These authors contribute this paper equally.}
\begin{abstract}
  Randomized controlled trials (RCTs) represent the paramount evidence of clinical medicine. Using machines to interpret the massive amount of RCTs has the potential of aiding clinical decision-making. We propose a RCT conclusion generation task from the PubMed 200k RCT sentence classification dataset to examine the effectiveness of sequence-to-sequence models on understanding RCTs. We first build a pointer-generator baseline model for conclusion generation. Then we fine-tune the state-of-the-art GPT-2 language model, which is pre-trained with general domain data, for this new medical domain task.
  Both automatic and human evaluation show that our GPT-2 fine-tuned models achieve improved quality and correctness in the generated conclusions compared to the baseline pointer-generator model. 
  Further inspection points out the limitations of this current approach and future directions to explore\footnote{The code is available at: \url{https://github.com/MiuLab/RCT-Gen}}.
  
\end{abstract}

\section{Introduction}

Randomized controlled trials (RCTs) are the most rigorous method to assess the effectiveness of treatments, such as surgical procedures and drugs, in clinical medicine \cite{Sibbald201}. A typical RCT often constitutes of two randomized groups of patients receiving either the ``intervention'' (new treatment) or ``control'' (conventional treatment). Then, a statistical analysis is done after the experiments to determine whether the intervention has a significant effect (i.e. actually making patients better or worse). The results from various RCTs contribute to the medical decisions made by physicians every day. However, analyzing these large amounts of data could be overwhelming for clinicians \cite{10.1001/jama.2011.619}. With the help of machine readers, we can alleviate the burden for providing correct information that contributes to critical clinical decisions.

In this work, we aim to evaluate the capabilities of deep learning models on understanding RCTs by generating the conclusions of RCT abstracts. We achieve this by transforming the PubMed 200k RCT abstract sentence classification dataset \cite{dernoncourt-lee-2017-pubmed} into a RCT conclusion generation task. Generating a \emph{correct} and \emph{coherent} conclusion requires the model to 1) identify the objectives of the trial, 2) understand the result and 3) generate succinct yet comprehensible texts. Therefore, this task can be a preliminary goal toward a more thorough understanding of clinical medicine literature. 

To tackle this task, we first build a pointer-generator model \cite{see-etal-2017-get} as the baseline. This model is widely used in abstractive summarization, which is similar to our conclusion generation task. We then leverage  the high quality text generation capability of the Open AI GPT-2 \cite{radford2019language} language model by fine-tuning the general domain GPT-2 model into a medical domain conclusion generator. 

Because the correctness of RCT understanding is essential for supporting clinical decisions and neural summarization models could inaccurately present facts from the source document, we incorporate human evaluation on the correctness and quality of the generated in addition to standard ROUGE score \cite{lin-2004-rouge} for automated summarization scoring. Evaluation results show the fine-tuned GPT-2 models score higher for both correctness and quality. However, there is still quite a large room for improvement both on the diversity and accuracy of the generated conclusions, providing a guidance for future research directions.

\section{Related Work}

The paper focuses on generating RCT conclusions, which is related to natural language generation.
We describe the related work below and emphasize the difference between the prior work and our work.
 In our proposed method, we exploit the state-of-the-art language model representations for understanding the complex medical literature, and related work is then briefly described below.

\subsection{Medical Natural Language Generation}

Several medical domain natural language generation tasks have been studied using machine learning models, including generating radiology reports from images \cite{jing-etal-2018-automatic,vaswani2017attention} and summarizing clinical reports \cite{zhang-etal-2018-learning-summarize, 10.1093/jamia/ocv032} or research literature \cite{cohan-etal-2018-discourse}.
Recently, \newcite{GULDEN2019114} studied extractive summarization on RCT descriptions.

Abstractive summarization, in which the model directly generates summaries from the source document, is closely related to our conclusion generation task. 
Most neural approaches for abstractive summarization were based on sequence-to-sequence recurrent neural networks (RNNs) with attention mechanisms \cite{devlin2019bert}.
The pointer-generator network \cite{see-etal-2017-get} combined a copy mechanism that directly copies words from the source document and a coverage mechanism to avoid repetition caused by the RNN-based decoder, achieving good performance by handling unseen information.
\newcite{devlin2019bert} further combined intra-encoder and intra-decoder attention with policy learning by using ROUGE-L score as the reward and improved the performance in terms of the evaluation metric.
\newcite{hsu-etal-2018-unified} combined an extractive model that provided attention on the sentence level and the pointer-generator architecture, and \newcite{cohan-etal-2018-discourse} also worked on abstractive summarization of long documents, including medical papers from the PubMed database, based on the pointer-generator network. 

However, our goal to generate conclusions is different from abstractive summarization in that summarization is to shorten the source document while preserving most of the important information, whereas our conclusion generation model gives one or two sentences describing the main outcome of the given trial.
Given the superior performance of pointer-generation networks from the above related summarization work, this paper uses the pointer-generation model as baseline and focuses on RCT conclusion generation instead of abstractive summarization.

\subsection{Contextualized Representations}

Recent advances of contextualized representation models, such as ELMo \cite{peters2018deep}, Open AI GPT \cite{radford2018language} and BERT \cite{devlin2019bert} achieved remarkable results across different natural language understanding tasks, such as question answering, entailment classification and named entity recognition. 
At the core of these models was language modeling, with either forward prediction used in GPT, bidirectional prediction used in ELMo, or masked prediction used by BERT. Variants of BERT also improved the performance of bio-medical natural language understanding tasks \cite{DBLP:journals/corr/abs-1906-04382,pugaliya2019pentagon}. \newcite{DBLP:journals/corr/abs-1906-05474} further proposed a new benchmark to evaluate the performance of contextualized models in the bio-medical domain.


Particularly, the Open AI GPT-2 model \cite{radford2019language} has demonstrated rudimentary zero-shot summarization capabilities with only language modeling training. Its forward prediction architecture made it suitable for autoregressive generation in a sequence-to-sequence task. 
Most benchmarks on contextualized representation were based on sequence classification tasks such as natural language inference and multiple choice question answering \cite{DBLP:journals/corr/abs-1804-07461,DBLP:journals/corr/abs-1906-05474}. 
Our work, on the other hand, focuses on exploring GPT-2's capability of generating goal-directed sentences in the medical domain.
Note that to our knowledge, this paper is the first attempt that investigates GPT-2 towards the medical document understanding and interpretation.

\begin{table*}[t!]
  \small
  \begin{center}
  \def\arraystretch{1.5}
  \begin{tabular}{|p{0.95\textwidth}|}
  \hline
  {\bf Source:}\newline
  \dblue{(BACKGROUND)}
  Varenicline is believed to work , in part , by reducing craving responses to smoking cues and by reducing general levels of craving ; however , these hypotheses have never been evaluated with craving assessed in the natural environments of treatment-seeking smokers . 
   
  \dblue{(OBJECTIVE)}
  Ecological momentary assessment procedures were used to assess the impact of varenicline on cue-specific and general craving in treatment-seeking smokers prior to quitting . 
   
  \dblue{(RESULTS)}
  During all phases , smoking cues elicited greater craving than neutral cues ; the magnitude of this effect declined after the first week . General craving declined across each phase of the study . Relative to the placebo condition , varenicline was associated with a greater decline in general craving over the drug manipulation phase . Varenicline did not significantly attenuate cue-specific craving during any phase of the study . \\
  \hhline{|=|}
  {\bf Target (\dgreen{True Negative}):}\newline Smoking cues delivered in the natural environment elicited strong craving responses in treatment-seeking smokers , but cue-specific craving was not affected by varenicline administered prior to the quit attempt . These findings suggest that the clinical efficacy of varenicline is not mediated by changes in cue-specific craving during the pre-quit period of treatment-seeking smokers . \\
  \hline
  {\bf Pointer-generator baseline model with $n = 1$ hint word (\dred{N/A}):}\newline \dblue{smoking} cues are associated with a greater craving in general , and may be associated with a greater decline in general craving and \\
  
  \hline
  {\bf Fine-tuned GPT-2 with $n = 0$ hint word (\dred{False Negative}):}\newline Varenicline did not reduce general craving in treatment-seeking smokers prior to quitting. \\ 
  \hline
  {\bf Fine-tuned GPT-2 with $n = 1$ hint word (\dgreen{True Negative}):}\newline \dblue{Smoking} cues are associated with greater general craving than neutral cues, and varenicline does not attenuate cue-specific craving.\\ 
  \hline
  \end{tabular}
  \end{center}
  \caption{An example of the GPT-2 $n = 0$ model generating a false negative conclusion (Varenicline did reduce general craving), while the GPT-2 $n = 1$ model generated a better true negative one. The ``(BACKGROUND)'', ``(OBJECTIVE)'' and ``(RESULTS)" tags denote the sentence classifications according to the original PubMed RCT dataset and are not included in the actual input of our conclusion generation task. }
  \label{tab:intro}
\end{table*}

\section{Task Formulation}

The PubMed 200k RCT dataset was originally constructed for sequential short text classification, with each sentence labeled as ``background'', ``objective'', ``methods'', ``results'' and ``conclusions''. 
We concatenated the ``background'', ``objective'' and ``results'' sections of each RCT paper abstract as the model input and the goal of the model is to generate the ``conclusions''. 
Table~\ref{tab:intro} illustrates the formulated task, where the generated conclusion needs to contain \emph{correct} information based on the experiments and should be \emph{concise}. 
After preprocessing, the number of abstracts in the training set is 189,035  and there are 2,479 conclusions used for validation.
The average source paragraph length is 170.1 words (6.0 sentences), and the average target conclusion length is 41.4 words (1.8 sentences) long.

\section{Models}

Language model pre-training has achieved a great success
among language understanding tasks with different model
architectures.
Because training language models requires a
large amount of text data, and it is relatively difficult to acquire a lot of RCT documents, this work focuses on first pre-training language models with the transformer architecture~\cite{vaswani2017attention} and then adapts the model to support the medical domain by fine-tuning.
The language model pre-training from general texts is described below.

\begin{figure}[t!]
    \centering
    \includegraphics[width=0.7\linewidth]{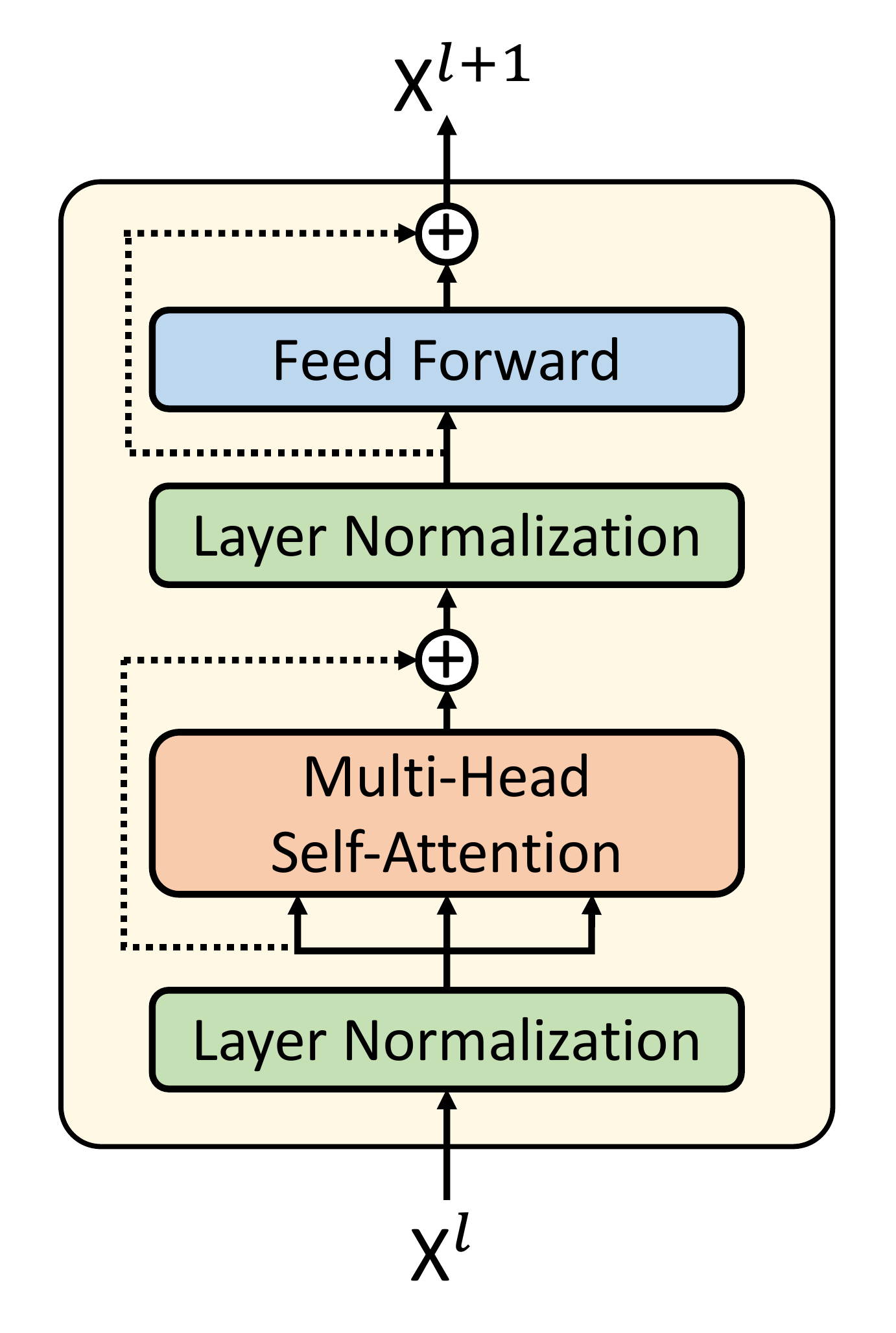}
    \caption{A modified transformer encoder block in the GPT-2 language model.}
    \label{fig:transformer}
\end{figure}

\subsection{Transformer Encoder in GPT-2}
We first introduce the transformer encoder \cite{vaswani2017attention} used as the backbone of the GPT-2 model. 
The transformer encoder is a stack of $N$ transformer encoder blocks, where the $l$-th block takes a sequence of hidden representations $X^l = \{X^l_1, \cdots , X^l_n\}$ as the input and outputs an encoded sequence $X^{l+1} = \{X^{l+1}_1, \cdots , X^{l+1}_n\}$.
A transformer encoder block consists of a multi-head self-attention layer and a position-wise fully connected feed-forward layer. A residual connection \cite{he2016deep} is employed around each of the two layers followed by layer normalization \cite{ba2016layer}. In GPT-2, however, the layer normalization step is moved to the front of the multi-head self-attention layers and the feed-forward layers.
An illustration of a GPT-2 transformer encoder block is presented in Figure \ref{fig:transformer}.
Each component is briefly described as follows.

\paragraph{Byte-Pair Encoding}

GPT-2 uses a special byte pair encoding (BPE) for input and output representations.
It can cover essentially all Unicode strings, which is useful in processing the medical texts due to the significant out-of-vocabulary problems such as distinct nomenclature and jargon. This special BPE prevents merging characters from different categories and preserves word-level segmentation properties with a space exception.

\paragraph{Positional Encoding}
Because the transformer model relies on a self-attention mechanism with no recurrence, the model is unaware of the sequential order of inputs.
To provide the model with positional information, positional encodings are applied to the input token embeddings
\begin{equation*}
    X^1_i = \text{embed}_{\text{token}}[w_i] + \text{embed}_{\text{pos}}[i],
\end{equation*}
where $w_i$ denotes the $i$-th input token, $\text{embed}_{\text{token}}$ and $\text{embed}_{\text{pos}}$ denote a learned token embedding matrix and a learned positional embedding matrix respectively.

\paragraph{Multi-Head Self-attention}
An attention function can be described as mapping a query to an output with a set of key-value pairs. The output is a weighted sum of values.
We denote queries, keys and values as $Q$, $K$ and $V$, respectively.
Following the original implementation \cite{vaswani2017attention}, a scaled dot-product attention is employed as the attention function. 
Hence, the output can be calculated as
\begin{equation*}
    \text{Attention}(Q, K, V) = \text{softmax}(\frac{QK^T}{\sqrt{d_k}})V,
\end{equation*}
where $d_k$ denotes the dimension of key vectors.

The idea of multi-head attention is to compute multiple independent attention heads in parallel, and then concatenate the results and project again. 
The multi-head self-attention in the $l$-th block can be calculated as
\begin{align*}
    & \text{MultiHead}(X^l) = \text{Concat}(\text{head}_1, \cdots, \text{head}_h)W^O, \\
    & \text{head}_i = \text{Attention}(X^lW^Q_i, X^lW^K_i, X^lW^V_i),
\end{align*}
where $X^l$ denotes the input sequence of the $l$-th block, $h$ denotes the number of heads, $W^Q_i$, $W^K_i$, $W^V_i$ and $W^O$ are parameter matrices.

\paragraph{Position-Wise Feed-Forward Layer}
The second sublayer in a block is a position-wise feed-forward layer, which is applied to each position separately and independently.
The output of this layer can be calculated as
\begin{equation*}
    \text{FFN}(x) = \max(0, x\cdot W_1 + b_1)W_2 + b_2,
\end{equation*}
where $W_1$ and $W_2$ are parameter matrices, $b_1$ and $b_2$ are parameter biases.

\paragraph{Residual Connection and Layer Normalization}
As shown in Figure~\ref{fig:transformer}, layer normalization is first applied on the input to the multi-head attention and feed-forward sublayers. The residual connection is then added around the two sublayers. The output of the $l$-th block can be calculated as
\begin{align*}
    H^{l} & = \text{MultiHead}(\text{LayerNorm}(X^l)) + X^l, \\
    X^{l+1}  & = \text{FFN}(\text{LayerNorm}(H^l)) + H^l.
\end{align*}

\subsection{GPT-2 Pre-Training}

The generative pre-training (GPT) via a language model objective is shown to be effective for learning representations that capture syntactic and semantic information without supervision \cite{peters2018deep,radford2018language,devlin2019bert}.
The GPT model proposed by \newcite{radford2018language} employs the transformer encoder with 12 encoder blocks. 
It is pre-trained on a large generic corpus that covers a wide range of topics.
The training objective is to minimize the negative log-likelihood:
\begin{equation*}
    \mathcal{L} = \sum_{t=1}^{T}{-\log P(w_t \mid w_{<t}, \theta)},
\end{equation*}
where $w_t$ denotes the $t$-th word in the sentence, $w_{<t}$ denotes all words prior to $w_t$, and $\theta$ are parameters of the transformer model.

To avoid seeing the future contexts, a masked self-attention is applied to the encoding process. 
In the masked self-attention, the attention function is modified into
\begin{equation*}
    \text{Attention}(Q, K, V) = \text{softmax}(\frac{QK^T}{\sqrt{d_k}} + M)V,
\end{equation*}
where $M$ is a matrix representing masks.
$M_{ij} = -\infty$ indicates that the $j$-th token has no contribution to the output of the $i$-th token, so it is essentially ``\emph{masked out}'' when encoding the $i$-th token.
Therefore, by setting $M_{ij} = -\infty$ for all $j > i$, we can calculate all outputs simultaneously without looking at future contexts.
It was pre-trained on the WebText dataset consisting of 40 GB high quality text crawled from internet sources.
We use the small version (12 layers and 117 M parameters) of the released GPT-2 models.

\begin{figure}[t!]
    \centering
    \subfloat[The pre-training stage]{\includegraphics[width=0.75\linewidth]{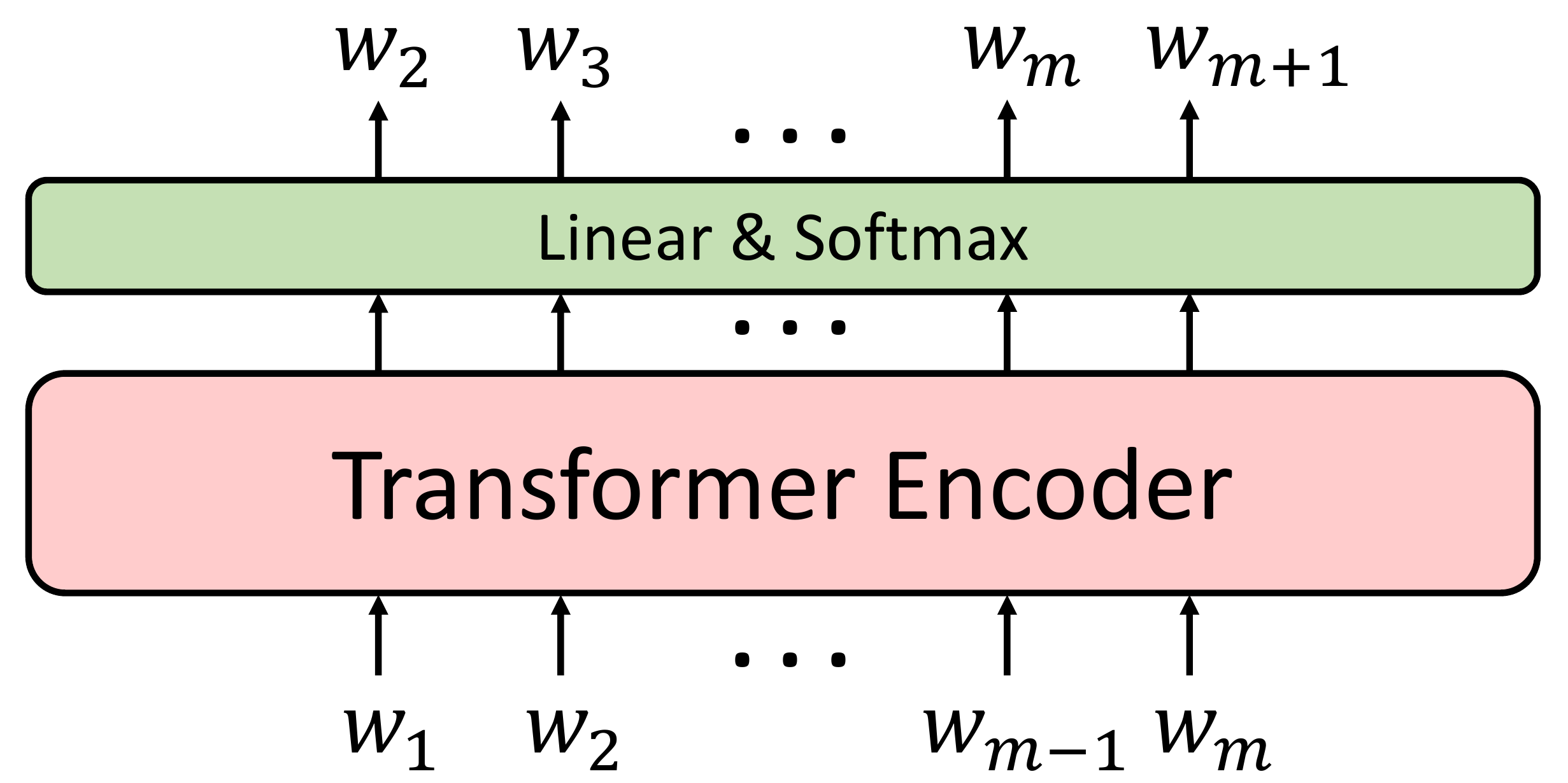}}\\
    \subfloat[The fine-tuning stage.]{\includegraphics[width=0.95\linewidth]{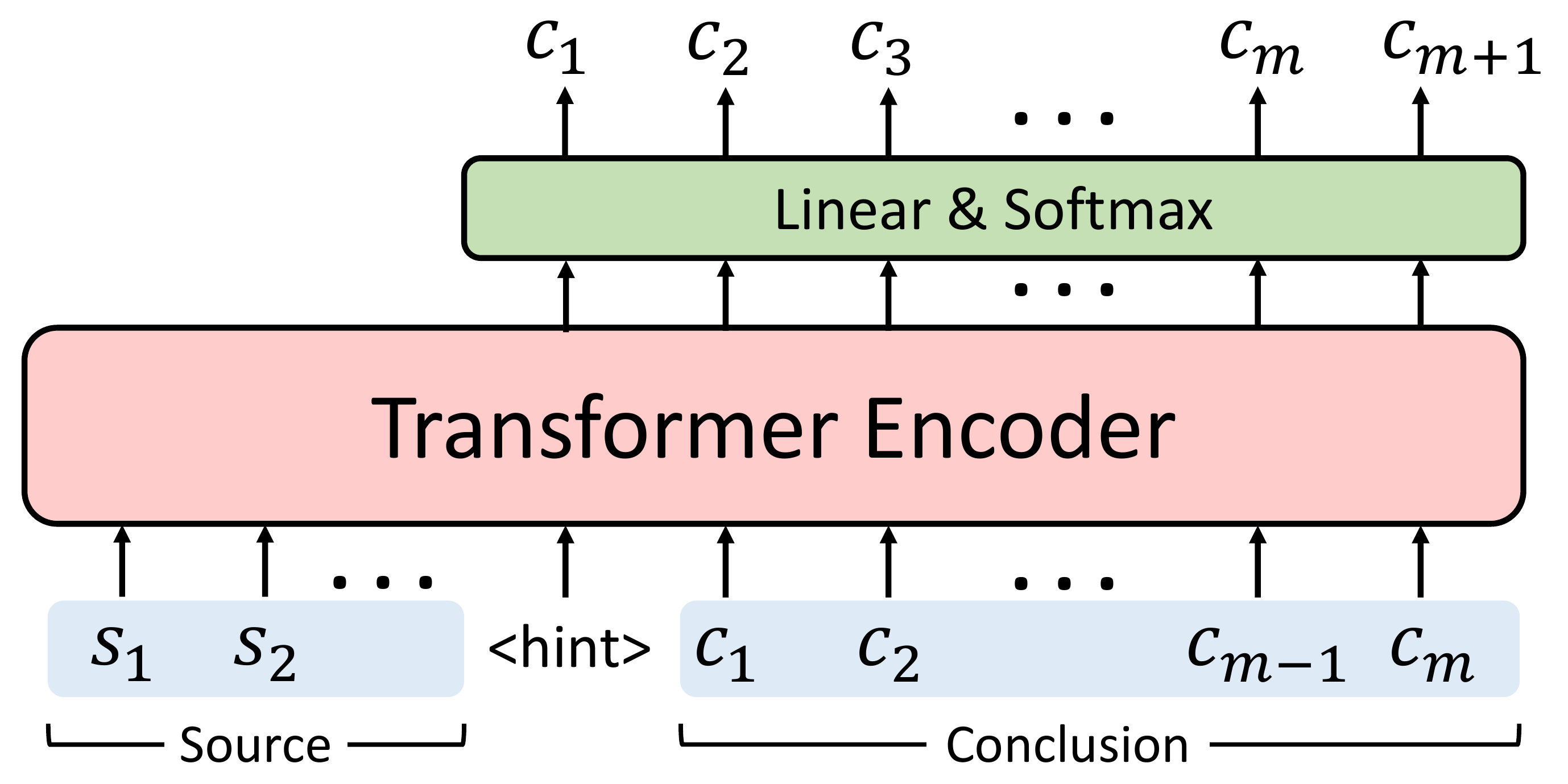}}
    \caption{Illustration of the two-stage method in the GPT model. The tag $\texttt{<hint>}$ denotes where hint word tokens are introduced during fine-tuning.}
    \label{fig:GPT}
\end{figure}

\subsection{GPT-2 Fine-Tuning}

After the model is pre-trained with a language model objective, it can be fine-tuned on downstream tasks with supervised data.
In our task, we adapt the GPT-2 to the target domain by fine-tuning using RCT data.
Figure~\ref{fig:GPT} illustrates the learning procedure.
By fine-tuning on the target data, the GPT-2 model may have the potential of understanding and generating medical texts. 

\begin{figure}[t!]
    \centering
    \includegraphics[width=0.9\linewidth]{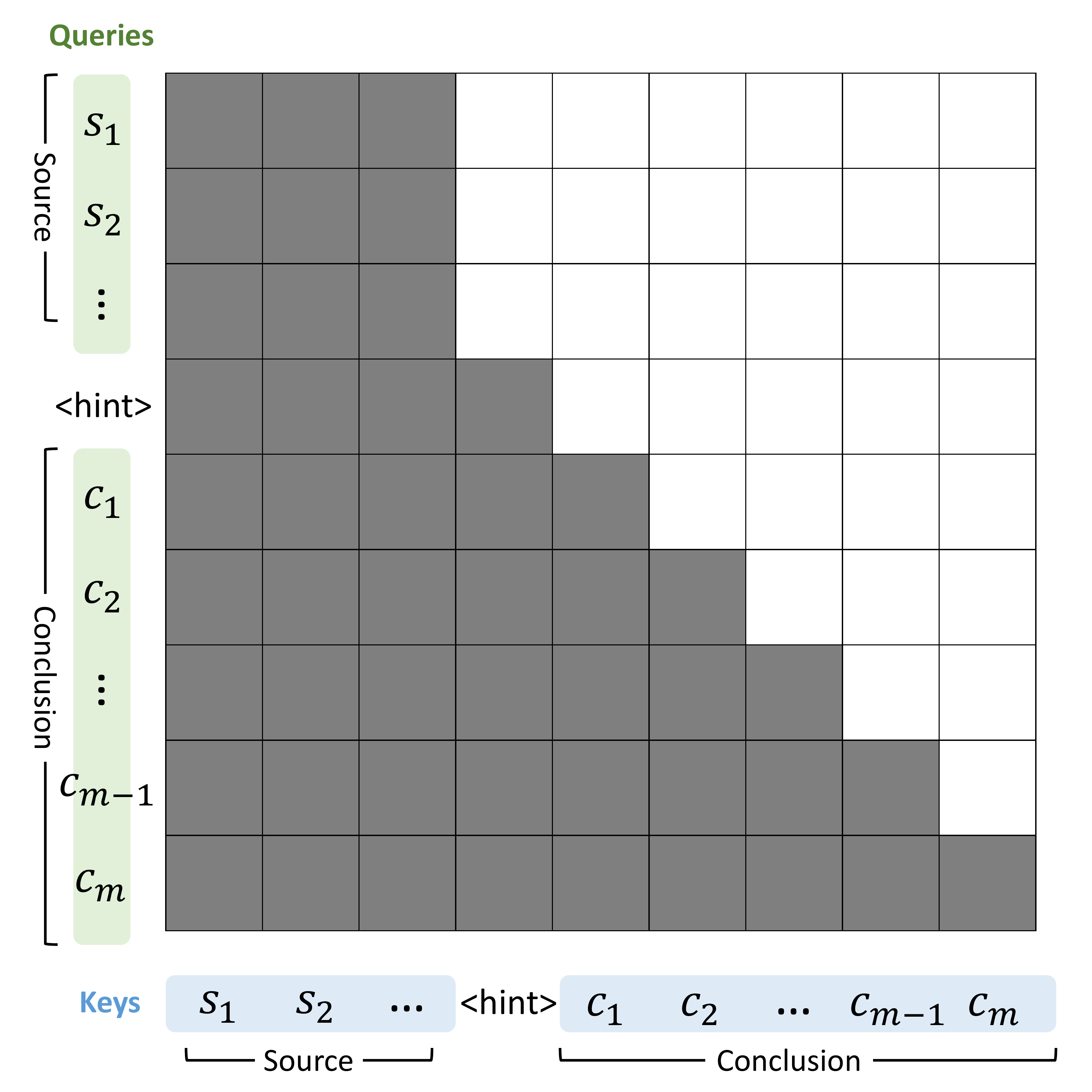}
    \caption{The attention mask used during fine-tuning. White cells denote $-\infty$ elements and grey cells denote $0$ in the mask matrix.}
    \label{fig:attn}
\end{figure}

In the fine-tuning stage, we modify the attention masking of the GPT-2 model so that source byte pairs are fully aware of the entire context of the source sentence, while the target byte pairs are aware of the entire source sentence plus the generated byte pairs that precede itself. That is, for context token pairs $(c_i, c_j) \in {c_1, \cdots, c_m}$, we set $M_{ij} = -\infty$ for all $j > i$, while for context and source token pairs $(c_i, s_j)$, where  $c_i \in {c_1, \cdots, c_m} $ and $s_j \in {s_1, \cdots, s_n} $, we set $M_{ij} = 0$. For all source token pairs $(s_i, s_j) \in {s_1, \cdots, s_n}$, we also set $M_{ij} = 0$. This setting is illustrated in Figure \ref{fig:attn}.

\section{Experiments}

Here we describe experimental details of the baseline pointer-generator model and the GPT-2 fine-tuned models.


\subsection{Experimental Setup}

The baseline model is a pointer-generator network \cite{see-etal-2017-get} with both copy and coverage mechanisms, and is trained with a coverage loss. We adopt the implementation of \newcite{zhang-etal-2018-learning-summarize}.
The vocabulary size is about 50,000, with uncased word embeddings pretrained from the PubMed RCT 200k training set and the abstracts from the PubMed dataset of long documents \cite{cohan-etal-2018-discourse}. We concatenate $n \in \{0, 1\}$ hint words following the source sentences, where the hint words are first $n$ words of the target conclusion. Our pointer-generator model uses beam search with beam size 5 to decode the final output conclusion.

In our GPT-2 models, we conduct conclusion generation using $n \in \{0, 1, 3, 5\}$ hint words. For $n = 0$, we append ``In conclusion , '' to the input.
Also, we perform data ablation study using only the ``results'' section as the model input. 
To address the memory constraint on our machines, we only train examples that are less than 500 byte pairs after encoding. Because GPT-2 model uses BPE for input and output, the generated conclusions are capitalized. Previous work showed that beam search did not help the generation quality of GPT-2 models \cite{DBLP:journals/corr/abs-1904-09751}, so we simply use greedy decoding to generate the conclusions .
Our GPT-2 model is fine-tuned with teacher forcing, using the SGD optimizer with learning rate of 0.001, momentum of 0.9 and the decay factor of 0.0005. Our model is based on a PyTorch implementation of GPT-2
\footnote{\url{https://github.com/huggingface/pytorch-transformers}}.



\subsection{Automatic Evaluation}

\begin{table}[t]
  \centering
  \setlength{\tabcolsep}{0.35em}
  \begin{tabular}{lccc}
    \toprule
    \bf System   & {\small \bf ROUGE-1} & {\small \bf ROUGE-2} & \small \bf ROUGE-L \\
    \midrule
    PGNet $n = 0$ & 27.11 & \phantom{0}7.61 & 21.87 \\
    PGNet $n = 1$ & 26.88 & \phantom{0}8.19 & 22.63 \\
    \midrule
    GPT-2 $n = 0$ & 30.33 & 11.34 & 25.14 \\
    GPT-2 $n = 1$ & \textbf{31.61} & \textbf{11.88} & \textbf{26.71} \\
    GPT-2 $n = 3$ & 29.94 & 11.55 & 25.85 \\
    GPT-2 $n = 5$ & 29.79 & 11.29 & 25.94 \\
    GPT-2 $res$ & 24.24 & \phantom{0}6.79 & 20.71 \\
    \bottomrule
  \end{tabular}
  \caption{ROUGE scores of the PGNet baseline models and the GPT-2 fine-tuned models on the development set. The GPT-2 $res$ were trained with the ``results'' section only. Addition of $n > 1$ hint words did not show further gains in ROUGE scores. }
\label{tab:rouge}
\end{table}

Table~\ref{tab:rouge} shows the best validation ROUGE scores of baselines and our models. 
Note that the hint words are not considered in score calculation and the output of all models are lower-cased.
The GPT-2 fine-tuned model significantly outperforms the pointer-generator (PGNet) baseline on all ROUGE scores, where the best performing model is GPT-2 with hint word $n = 1$, demonstrating the effectiveness of generating good conclusion in our model.
However, more hint words do not bring additional gain in ROUGE scores, probably because more constraints hinder the GPT-2 model to explore potentially good conclusions.
Moreover, the ablation result shows the significant drop in all ROUGE scores, indicating the importance of including the ``background'' and ``objective'' sections in the input for better content understanding.

\begin{table}[t!]
  \centering
  \setlength{\tabcolsep}{0.35em}
  \begin{tabular}{lcccccc}
    \toprule
    \bf System   & \bf TP & \bf TN & \bf FP & \bf FN & \bf N/A & \bf Acc.\\
    \midrule
    PGnet $n = 1$ & 15 & 3 & 5 & 3 & 24 & 36\%\\
    \midrule
    GPT-2 $n = 0$ & 24 & 3 & 4 & 5 & 14 & 54\%\\
    GPT-2 $n = 1$ & \textbf{26} & \textbf{6} & \textbf{5} & \textbf{3} & \textbf{10} & \bf 64\%\\
    \midrule
    Target & 32 & 11 & 0 & 0 & 7 & 86\% \\
    \bottomrule
  \end{tabular}
  \caption{Human evaluation results for text understanding on the annotation questions of 50 randomly selected source documents. Note that some source documents which don't fit into the binary paradigm of positive or negative results are classified as N/A. TP: True Positive; TN: True Negative; FP: False Positive; FN: False Negative; N/A: Not Applicable.}
\label{tab:annotation}
\end{table}

\subsection{Human Evaluation}

We recruited 10 medical students with prior training in bio-statistics and epidemiology to annotate and rate the generated conclusions. 
Our questionnaire contains two types of questions: the annotation question and the rating question. 
\begin{itemize}
\item \textbf{Annotation}: A annotation question contains a source document and four conclusions, namely the target conclusion written by human, the GPT-2 $n = 0$, the GPT-2 $n = 1$ and the PGnet generated conclusions. 
The raters are asked to classify each generated conclusions as either true positive, true negative, false positive, false negative or not applicable. We define true / false as whether the generated conclusion corresponds to the given document, and positive / negative as whether the intervention studied has a statistically significant effect, regardless of the effect being favourable or detrimental to the patients.
This is to explicitly examine whether the generated conclusion is precise in terms of RCT content understanding.
\item \textbf{Rating}: Rating questions use the same set-up except the question is a 5 point Likert scale for \emph{correctness}, \emph{quality} and \emph{overall} impression. Each rater is given 5 annotation questions and 5 rating questions, with each source document randomly chosen from the validation set. 
This is to judge the generated conclusions both regarding to and regardless of the source document.
\end{itemize}

To mitigate bias, we do not inform which conclusion was generated or written by human, and the conclusions are lower-cased and randomly ordered in each question for fair comparison.

Table~\ref{tab:annotation} presents the results from the annotation question, where the number of true positive and true negative generations from the GPT-2 fine-tuned models increase when compared to the PGNet baseline. 
It is clear that the proposed GPT-2 achieves better performance in terms of accuracy (the ratio of true samples).
We also include the performance of human-written conclusions in the last row, which serves as the upper bound of this task.
However, there is still a gap between human-written conclusions and the generated ones.

In the rating questions depicted in Table~\ref{tab:rating}, the human written conclusions obtain a score nearly 4 out of 5 on all three dimensions. The GPT-2 models have comparable scores in overall impression, both scoring around 3.5 out of 5. 
The most significant improvement of the GPT-2 generated conclusions is the text quality, with the correctness improvement to a lesser extent. The correctness of GPT-2 $n=1$ is slightly better than that of GPT-2 $n=0$ in the annotation question, yet in the rating question, GPT-2 $n=0$ has a higher averaged score.
In sum, the human evaluation results demonstrate that our models significantly outperform the baseline pointer generator and tell that the proposed RCT conclusion generation task is not the same as typical summarization task, so deep text understanding is required for better performance.

\begin{table}[t!]
  \centering
  \setlength{\tabcolsep}{0.35em}
  \begin{tabular}{lcccc}
    \toprule
    \bf System   & \bf Correctness & \bf Quality & \bf Overall \\
    \midrule
    PGnet $n = 1$ & 3.02 & 2.86 & 2.86 \\
    \midrule
    GPT-2 $n = 0$ & \textbf{3.42} & 3.66 & \textbf{3.52} \\
    GPT-2 $n = 1$ & 3.30 & \textbf{3.94} & 3.50 \\
    \midrule
    Target & 3.92 & 4.08 & 3.98  \\
    \bottomrule
  \end{tabular}
  \caption{Human evaluation results for generation quality on the rating questions.}
\label{tab:rating}
\end{table}

\begin{table*}[t!]
  \small
  \begin{center}
  \def\arraystretch{1.5}
  \begin{tabular}{|p{0.95\textwidth}|}
  \hline
  {\bf Source:} Proton pump inhibitor ( PPI ) therapy is considered as the first choice for treatment of non-erosive reflux disease ( NERD ) . However , NERD is less sensitive to PPIs than erosive gastroesophageal reflux disease ( GERD ) and the differences between PPIs and H2 receptor antagonists are less evident in NERD than in erosive GERD . Since gastric acid secretion is lower in the Japanese population than in Western populations , we aimed to investigate whether PPI therapy is really necessary for NERD patients in Japan . Both roxatidine and omeprazole significantly improved the heartburn score at 4 and 8 weeks . The clinical response rates did not differ between roxatidine and omeprazole . Both roxatidine and omeprazole significantly relieved not only reflux but also abdominal pain and indigestion . The degrees of improvement did not differ between the two groups . \\
  \hhline{|=|}
  {\bf Target (\dgreen{True Positive}):} Roxatidine relieved the symptoms of NERD patients with similar effectiveness to omeprazole . Therefore , roxatidine may be a good choice for NERD treatment .\\
  \hline
  {\bf GPT-2 $n = 0$ (\dgreen{True Positive}):} Both roxatidine and omeprazole significantly improved the heartburn score at 4 and 8 weeks. \\
  \hline
  {\bf GPT-2 $n = 1$ (\dgreen{True Positive}):} \dblue{Roxatidine} and omeprazole are effective in relieving symptoms of NERD in Japanese patients. \\ 
  \hline
  \end{tabular}
  \end{center}
  \caption{An example of GPT-2 models generating true positive conclusions.}
  \label{tab:sup1}
\end{table*}

\begin{table*}[t!]
  \small
  \begin{center}
  \def\arraystretch{1.5}
  \begin{tabular}{|p{0.95\textwidth}|}
  \hline
  {\bf Source:} To evaluate the efficacy of oxcarbazepine ( OXC ) in the treatment of agitation and aggression in patients with Alzheimer 's disease , vascular dementia or both . In total , 103 institutionalized patients at 35 sites were randomized to the trial . After 8 weeks , no statistically significant differences were found between the 2 groups for all outcomes . A trend was observed in favor of the OXC group in the reduction in the scores on the BARS ( p = 0.07 ) . \\
  \hhline{|=|}
  {\bf Target (\dgreen{True Negative}):} This study found no significant effect of OXC in treatment of agitation and aggression in patients with dementia . \\
  \hline
  {\bf GPT-2 $n = 0$ (\dgreen{True Negative}):} OXC was not effective in the treatment of agitation and aggression in patients with Alzheimer's disease, vascular dementia or both. \\
  \hline
  {\bf GPT-2 $n = 1$ (\dred{False Positive}):} \dblue{This} study suggests that OXC is effective in the treatment of agitation and aggression in patients with Alzheimer's disease. \\ 
  \hline
  \end{tabular}
  \end{center}
  \caption{An example of GPT-2 $n = 0$ model generating a true negative conclusion, while the GPT-2 $n = 1$ model generated a false positive one.}
  \label{tab:sup2}
\end{table*}

\begin{table*}[t!]
  \small
  \begin{center}
  \def\arraystretch{1.5}
  \begin{tabular}{|p{0.95\textwidth}|}
  \hline
  {\bf Source:} Atrial fibrillation ( AF ) is the most common complication following coronary artery bypass graft ( CABG ) . The mechanism of AF after CABG is not well defined ; however , it is suggested that endogenous adenosine , released in response to tissue hypoxia , may play a mechanistic role in these arrhythmias . The purpose of this study was to examine whether intravenous theophylline , via adenosine A1 receptor antagonism , would correct or modify new-onset early ( \textless 48 h post CABG ) atrial fibrillation in patients post CABG , and thereby implicate endogenous adenosine as an inciting agent . Thirty patients comprised the study group . In Group A , 8 of the 15 patients ( 53 \% ) converted from AF to sinus rhythm within 15 min of theophylline administration . One patient who converted to sinus rhythm 20 min after theophylline administration was accepted as showing a negative response . In the placebo-treated group , no patient converted to sinus rhythm within 15 min ( p \textless 0.007 compared with Group A ) . \\
  \hhline{|=|}
  {\bf Target (\dgreen{True Positive}):} The mechanism of AF after CABG is not well defined and is probably multifactorial . However , this study demonstrated that antagonism of the adenosine A1 receptor can promptly convert many of these patients back to sinus rhythm , and thereby implicates endogenously released adenosine in a mechanistic role for inciting early ( \textless 48 h ) post-CABG AF . \\
  \hline
  {\bf GPT-2 $n = 0$ (\dred{False Negative}):} Intravenous theophylline, via adenosine A1 receptor antagonism, did not improve early AF in patients post CABG. \\
  \hline
  {\bf GPT-2 $n = 1$ (\dgreen{True Positive}):} \dblue{The} results of this study suggest that intravenous theophylline, via adenosine A1 receptor antagonism, may correct or modify early AF in patients post CABG. \\ 
  \hline
  \end{tabular}
  \end{center}
  \caption{An example of GPT-2 $n = 0$ model generating a false negative conclusion, while the GPT-2 $n = 1$ model generated a true positive one.}
  \label{tab:sup4}
\end{table*}

\section{Discussion}

From the human evaluation results and our empirical inspection, we discover two major problems concerning the quality of the generated conclusions from GPT-2 models. 
First, there is some repetition in the generated conclusions, which impair the quality of generated text, though not as common in that of RNN-based models.
We suggest additional weighted decoding or coverage mechanisms to avoid such problems. 
Second, the GPT-2 generated conclusions are significantly shorter than the targets. The average length generated by GPT-2 $n=0$ and GPT-2 $n=1$ are 19.4 and 21.0, while that of human written conclusions is 41.4. This could be caused by the limitation of greedy decoding, but the examples generated by PGnet, which applies beam search, only gives an average length of 22.6. This suggests investigation of additional measures to enrich and lengthen the generated conclusions in future work. 

Another important issue is the correctness of the generation model. 
The GPT-2 models are able to identify simple patterns and generate conclusions with the correct relationship. 
However, errors occur when the study design becomes more complicated or the outcomes are complex. 
Therefore, future work should aim at enhancing the language understanding capabilities of generation models.
Methods such as pre-training the GPT-2 models with medical domain literature or using external background knowledge might fill the missing gap in the correctness performance. This is very crucial regarding to our RCT understanding task and other tasks that require precise and reliable language generation. 


Here we select 3 examples to better illustrate our evaluation methods and the discussed limitations of the current models. The example in Table~\ref{tab:sup1} show two successful generations from the GPT-2 models. 
Table~\ref{tab:sup2} shows a false positive example by the GPT-2 $n=1$ model. On the other hand, a false negative example generated by the GPT-2 $n=0$ can be seen in Table~\ref{tab:sup4}. 
The generated conclusions in Table~\ref{tab:sup4} is also much shorter than the target conclusion written by human. Other factors that could cause this issue may include that the human authors mention information not included in the preceding source document, additional comments on the results and background knowledge and they paraphrase the same concept in different ways. 

Given the above results, this paper opens a new research direction by formulating the RCT conclusion generation task and investigates the potential of language generation models towards better understanding of medical documents.

\section{Conclusion and Future Work}

This work introduces the RCT paper conclusion generation task as a stepping stone to the automatic understanding of clinical research literature. 
Our results show the general domain pre-trained GPT-2 language model can be fine-tuned to generate medical domain conclusions.
The evaluation results show improvements regarding to both quality and correctness in conclusions generated by the fine-tuned GPT-2 model compared to the pointer-generator summarization model. 
Further study is needed to enhance the generation quality by reducing repetition errors and increasing the generation length, and to improve the correctness through better language understanding for practical clinical scenarios.

Beyond generating conclusions for RCT papers, generative language models in the medical domain with improved correctness and quality can open up new opportunities to tasks that require profound domain knowledge. For example, automatic generation of systemic review and meta-analysis articles.

\section*{Acknowledgements}
We would like to thank reviewers for their insightful comments on the paper.
This work was financially supported from the Young Scholar Fellowship Program by Ministry of Science and Technology (MOST) in Taiwan, under Grant 108-2636-E002-003 and 108-2634-F-002-015.

\bibliography{emnlp-ijcnlp-2019}
\bibliographystyle{acl_natbib}

\end{document}